\documentclass[10pt,twocolumn,letterpaper]{article}
\usepackage[table]{xcolor}

\usepackage[pagenumbers]{iccv} 
\usepackage{amsmath,amsfonts,bm}

\def\eqref#1{equation~\ref{#1}}

\def\1{\bm{1}}

\def\rvx{{\mathbf{x}}}

\DeclareMathAlphabet{\mathsfit}{\encodingdefault}{\sfdefault}{m}{sl}
\SetMathAlphabet{\mathsfit}{bold}{\encodingdefault}{\sfdefault}{bx}{n}

\newcommand{\E}{\mathbb{E}}

\definecolor{iccvblue}{rgb}{0.21,0.49,0.74}
\newcommand{\spara}[1]{\noindent\textbf{#1.}}
\usepackage[pagebackref,breaklinks,colorlinks,allcolors=iccvblue]{hyperref}
\newtheorem{lemma}{Lemma}[section]

\usepackage{multicol}
\usepackage{multirow}

\newcommand{\shortname}{JDM\xspace}

\title{Adding Additional Control to \\ One-Step Diffusion with Joint Distribution Matching}

\author{Yihong Luo $^{1}$\textsuperscript{*} \hspace{6mm} 
Tianyang Hu $^{2}$ \hspace{6mm}
Yifan Song $^{3}$ \hspace{6mm} \\
Jiacheng Sun $^{2}$ \hspace{6mm} 
Zhenguo Li $^{2}$ \hspace{6mm} 
Jing Tang $^{3,1}$\textsuperscript{†}  \hspace{6mm}  \\ \vspace{-3mm} \\
$^1$ HKUST \hspace{5mm} $^2$ Huawei Noah’s Ark Lab \hspace{5mm} $^3$ HKUST (GZ) \hspace{5mm} \\
\\
}

\begin{document}

\twocolumn[{%
 \renewcommand\twocolumn[1][]{#1}%
 \maketitle
 \centering
 \includegraphics[width=1\textwidth]{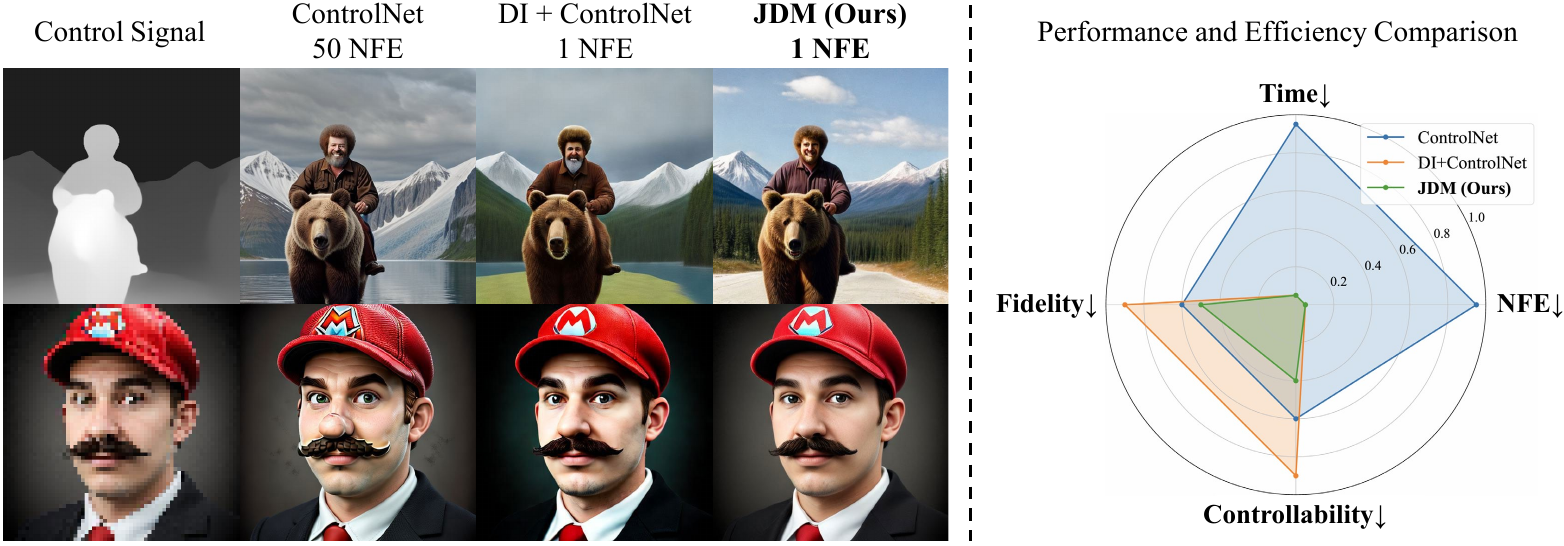}
 \captionof{figure}{Visual comparison of different strategies of adding controls. The compared baselines include 1) the diffusion with integrated standard ControlNet (denoted as ControlNet), and 2) the integration of pre-trained standard ControlNet with Diff-Instruct's pre-trained one-step generator (denoted as DI + ControlNet). Notably, our method not only maintains computational efficiency but also surpasses the visual quality achieved by the standard ControlNet approach. While the standard ControlNet approach relies heavily on high Classifier-Free Guidance (CFG) to achieve high-quality generation, this dependency might introduce unwanted artifacts in the final samples.
    \label{fig:teaser}
   }
   \vspace{2mm}
}]
\footnotetext[1]{Work was partly done during an internship at Huawei Noah’s Ark Lab.}
\footnotetext[2]{Corresponding Author.}

\begin{abstract}
While diffusion distillation has enabled one-step generation through methods like Diff-Instruct~\citep{luo2023diff} and Variational Score Distillation~\citep{wang2023prolificdreamer}, adapting distilled models to emerging \textit{new controls} -- such as novel structural constraints or latest user preferences -- remains challenging. 
Conventional approaches typically require modifying the base diffusion model and redistilling it -- a process that is both computationally intensive and time-consuming. 
To address these challenges, we introduce \textbf{J}oint \textbf{D}istribution \textbf{M}atching (\textbf{JDM}), a novel approach that minimizes the reverse KL divergence between image-condition joint distributions. By deriving a tractable upper bound, JDM decouples fidelity learning from condition learning. This asymmetric distillation scheme enables our one-step student to handle controls unknown to the teacher model and facilitates improved classifier-free guidance (CFG) usage and seamless integration of human feedback learning (HFL). 
Experimental results demonstrate that JDM surpasses baseline methods such as multi-step ControlNet~\citep{zhang2023adding} by mere one-step in most cases, while achieving state-of-the-art performance in one-step text-to-image synthesis through improved usage of CFG or HFL integration.
\end{abstract}

\vspace{-5mm}
\section{Introduction}
\vspace{-2mm}
\label{sec:intro}
Diffusion models (DMs)~\cite{ho2020denoising,sohl2015deep} have significantly advanced generative modeling, particularly in text-to-image synthesis, by producing high-quality and diverse images in a controllable manner~\cite{zhang2023adding}. 
However, their practical deployment is often hindered by the inefficiency of the sampling process, which usually takes tens to hundreds of Network Function Evaluations (NFEs). 
Thanks to the recent progress in diffusion distillation, the sampling efficiency has been greatly enhanced, and photo-realistic images can be generated with as few as 1 NFE~\cite{luo2023diff,yoso,yin2023one}.
With huge computational savings in model serving, diffusion distillation of the base DM has become the standard procedure. 

As artificial intelligence-generated content (AIGC) applications continue to evolve, new scenarios demand models to adapt to novel conditions and controls. 
These conditions encompass structural constraints, semantic guidelines, and external factors such as user preferences and additional sensory inputs. 
The conventional approach to integrating such controls into diffusion models entails modifying the base model and subsequently performing diffusion distillation for one-step student~\cite{song2024sdxs} --- a process that is both computationally expensive and time-intensive. 
A more efficient alternative would be to extend the distillation pipeline to accommodate new controls directly, bypassing the need for extensive retraining.
    
Existing work on this important matter is scarce.
On one hand, learning additional control for one-step students remains largely unexplored.
Doing so for the base DM typically relies on pre-trained ControlNet models optimized through denoising score matching (DSM) to obtain new controllability \citep{zhang2023adding,xiao2024ccm}. 
However, extending ControlNet for one-step generation incurs significant limitations, such as degraded fine-grained control and suboptimal sample quality, underscoring the need for new learning paradigms that better integrate control mechanisms into one-step generators.
On the other hand, incorporating additional control during diffusion distillation is also challenging.  
Current diffusion distillation methodologies for one-step generation predominantly focus on distilling a student model that replicates the capabilities of the teacher diffusion model~\cite{yin2023one,luo2023diff,song2024sdxs}, without investigating how to extend the student’s abilities \textit{beyond} those of the teacher. 
This limitation is particularly relevant when adding novel controls that the original diffusion model was not designed to handle.

To address these challenges, we propose a novel approach termed JDM, that minimizes the reverse Kullback-Leibler (KL) divergence between the image-condition joint distributions. 
We derive a tractable upper bound for this divergence, which effectively decouples fidelity learning from condition learning. The \textit{asymmetrical} nature of our objective enables us to obtain a one-step student that can handle controls unknown to the teacher diffusion model. 
Moreover, this decoupling mechanism not only facilitates improved usage of classifier-free guidance (CFG), but also enables the seamless integration of human feedback learning (HFL) into the training process. Consequently, our method enhances both the controllability and quality of generated images, providing a more flexible and efficient framework for one-step diffusion generation.

Extensive experiments demonstrate the superiority of our proposed \shortname. For controllable generation tasks, our one-step approach achieves better performance than multi-step controllable DM (50 NFE), with lower FID scores in average (14.58 vs 15.21) and better controllability measured by consistency scores (improves 24\% in average). Besides, in text-to-image generation, by incorporating either human feedback learning or improved usage of CFG, our method establishes new state-of-the-art (SOTA) performance among one-step approaches. Specifically, our variant with better CFG achieves CLIP scores of 33.97, clearly outperforming the multi-step DM's 33.03.  

\vspace{-2mm}
\section{Background}
\vspace{-1mm}
\spara{Diffusion models (DMs)} DMs~\citep{sohl2015deep, ho2020denoising} establish a forward diffusion process that progressively introduces Gaussian noise into the data over $T$ steps: $q(\rvx_t|\rvx) \triangleq \mathcal{N}(\rvx_t; \alpha_t\rvx, \sigma_t^2\textbf{I})$, where $\alpha_t$ and $\sigma_t$ are hyper-parameters that control the diffused schedule. The diffused samples can be directly calculated as $\rvx_t = \alpha_t\rvx + \sigma_t \epsilon$, with $\epsilon \sim \mathcal{N}(\mathbf{0},\mathbf{I})$. The diffusion network $\epsilon_\phi$ is then trained to perform denoising by minimizing: $\E_{\rvx,\epsilon,t} || \epsilon_\phi(\rvx_t,t) - \epsilon||_2^2$. Once trained, the score of the diffused samples $\rvx_t$ can be estimated using:
\begin{equation}
\small
    \nabla_{\rvx_t} \log p(\rvx_t) \approx \nabla_{\rvx_t} \log p_\phi(\rvx_t) = s_\phi(\rvx_t,t) = - \tfrac{\epsilon_\phi(\rvx_t, t)}{\sigma_t},
\end{equation}
With the estimated score, sampling from DMs can be achieved by solving the corresponding diffusion stochastic differential equations (SDEs) or probability flow ordinary differential equations (PF-ODEs) with multiple steps.

\spara{ControlNet} Among other methods~\cite{mou2023t2i,bansal2024universal}, ControlNet~\cite{zhang2023adding} is the most promising method in adding additional Control to DMs. Given a pretrained diffusion model $\epsilon_\phi$, a ControlNet parameterized by $\beta$ can be trained by minimizing the denoising loss $L(\beta)$ for injecting additional controls, where $L(\beta) = \E_{\rvx,\epsilon,t}|| \epsilon - \epsilon_{\phi,\beta}(\rvx_t,c) ||_2^2$.

\spara{Diffusion distillation} 
Despite the active progress in training-free accelerated sampling of DMs~\cite{lu2023dpm, zhao2023unipc, xue2024accelerating, si2024freeu, ma2024surprising}, diffusion distillation is dispensable for satisfactory few-step generation.
Diffusion distillation typically follows two main appealing approaches:
: 1) Trajectory distillation~\cite{luhman2021knowledge, on_distill, luhman2021knowledge, salimans2021progressive, song2023consistency, song2024improved, yan2024perflow}, which attempts to replicate the teacher model's ODE trajectories on  instances level. These methods face challenges in difficult instance-level matching;
2) Distribution matching via score distillation~\cite{yin2023one, luo2023diff, sid},
which aims to replicate the teacher model on distribution level using distribution divergence metrics.

\spara{Score Distillation}  Diff-Instruct (DI)~\cite{luo2023diff} and Variational Score Distillation (VSD)~\cite{wang2023prolificdreamer} train a conditional one-step student by minimizing the reverse integral KL divergence:
\begin{equation}
\small
\E_t \lambda_t\mathrm{KL}(p_\theta(\rvx_t|c) || p(\rvx_t|c)), \ \lambda_t >0
\end{equation}
where $p_\theta(\rvx_t|c)\triangleq \int p_\theta(\rvx|c)q(\rvx_t|\rvx)d\rvx$ is the distribution of diffused sample. A student trained in this way can effectively replicate the teacher’s capabilities while enabling one-step generation. However, training a conditional student requires a two-step process: first, a conditional teacher must be trained, followed by distillation to transfer its knowledge to the student. This sequential approach renders the introduction of new control guidance inefficient.

\spara{Additional Controls for One-step Diffusion} 
Distilling one-step generator by score distillation has been well explored~\cite{luo2023diff,yin2023one,sid}, however, how to distill one-step generator with additional controls has not been well explored. 
CCM~\cite{xiao2024ccm} explores integrating consistency training with ControlNet, showing reasonable performance with four steps, while our work can surpass the standard ControlNet with mere one step in most cases.
The success of previous work~\cite{luo2023diff,yin2023one,dmd2} in score distillation relies on initializing one-step students with the teacher. SDXS~\cite{song2024sdxs} explored learning one-step generator with control via score distillation, however, their teacher and fake score are required to have a ControlNet that supports the injected condition. 
In contrast, we minimize a tractable upper bound of joint KL divergence. This approach enables an asymmetric formulation between the teacher and student, where our student is partially initialized by the teacher, supplemented with an additional ControlNet. Our strong empirical results demonstrate that it is possible to train a one-step student through score distillation, allowing it to understand conditions that the teacher does not.

\spara{Human Preference Alignment For One-Step Diffusion} Since our framework supports universal guidance, we also explore integrating the reward model as additional guidance into training one-step generators for aligning with human preference. Although there are many works that try to align diffusion models with human preferences~\cite{dai2023emu,podell2023sdxl,prabhudesai2023aligning,clark2023directly,lee2023aligning,fan2024reinforcement,black2023training,ye2024schedule}, we note that how to align a one-step generator with human preference is not well explored. The existing few works~\cite{ren2024hypersd,luo2025diffinstruct} directly maximize the reward of generated images, which can lead to obvious artifacts (see \cref{fig:rlhf_ablation}). In contrast, our method integrates preference learning with modeling $\log p(c|\rvx_t)$, suffering less from artifacts and providing better visual quality.
Besides, our frameworks decouple the condition and fidelity learning. This provides flexibility in applying CFG by an additional teacher model, showing better performance in distilling a one-step generator.

\vspace{-2mm}
\section{Methodology}
\vspace{-1mm}
\begin{figure}
    \centering
    \includegraphics[width=1\linewidth]{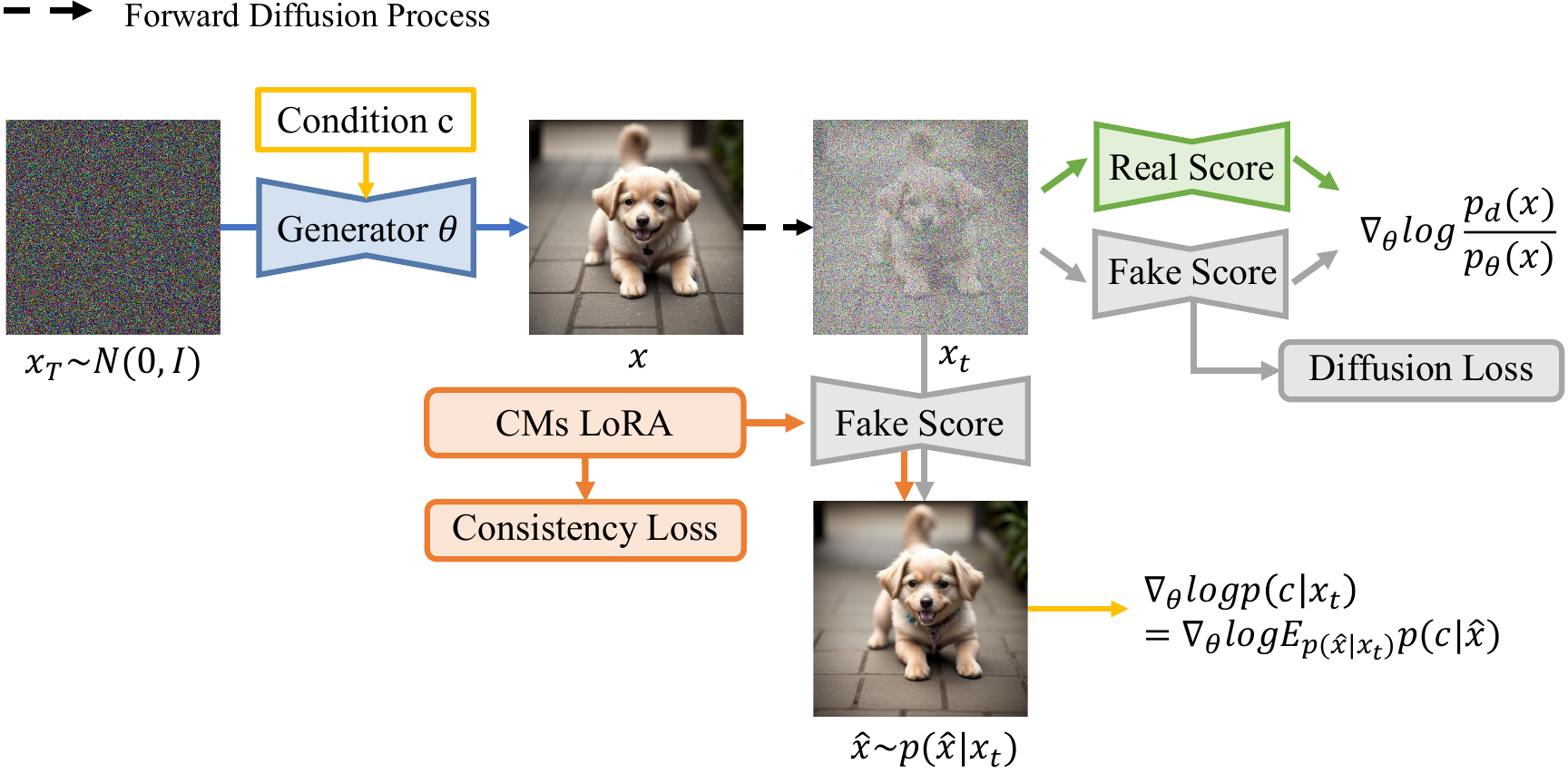}
    \vspace{-5mm}
    \caption{The framework description of our proposed \textbf{JDM}.}
    \vspace{-5mm}
    \label{fig:enter-label}
\end{figure}

\spara{Problem Setup} Consider a pre-trained DM with a multi-level score network $\bm{s}_\phi(\rvx_t, t) = \nabla_{\rvx_t} \log p_\phi(\rvx_t,t) \approx \nabla_{\rvx_t} \log p_t(\rvx_t)$, where $p_t(\rvx_t)$ represents the marginal diffused distributions at time $t$.
We assume that this pre-trained model provides a high-quality approximation of the data distribution, such that $p(\rvx_0) \approx p_d$. 
Additionally, we want to implement a new control $c$ given by a conditional discriminative model $\log p(c|\rvx)$. 
Our objective is to train a one-step generator that incorporates additional control $c$ through marginal diffusion. In essence, we aim to develop an algorithm that enables a student model to acquire new capabilities beyond those of the teacher diffusion model.

In order to directly inject new controls in learning one-step student, we propose minimizing the joint reverse KL divergence between $p_\theta(\rvx_t,c) \triangleq p_\theta(\rvx_t|c)p(c)$ and $p(\rvx, c) \triangleq p(c|\rvx_t)p_\phi(\rvx_t)$, i.e.,
\begin{equation}
\small
\label{eq:joint_kl}
\begin{aligned}
    & \E_t \lambda_t \mathrm{KL}(p_\theta(\rvx_t,c) || p(\rvx_t, c)) \\
    & = \E_t\lambda_t\mathrm{KL}(p_\theta(\rvx_t|c)p(c) || p(c|\rvx_t)p_\phi(\rvx_t))  \\
    & =  \lambda_t\E_{p_\theta(\rvx_t|c)p(c),t} [- \log p(c|\rvx_t)p_\phi(\rvx_t)  + \log p_\theta(\rvx_t,c)],
\end{aligned}
\end{equation}
where $\lambda_t>0$, $p(c)$ is a known fixed distribution. 
Notably, the joint KL divergence exhibits asymmetry between the target and student distributions. The target joint distribution factorizes into the marginal distribution $p_\phi(\rvx)$ and the conditional distribution $p(c|\rvx_t)$, with the latter being accessible through a discriminative model. This formulation enables us to distill a one-step conditional generative student that handles condition unknown to teacher generative model.

However minimizing the KL divergence in \cref{eq:joint_kl} still requires access to the gradient of the conditional student. Luckily, we can access its tractable upper bound as follows.
\begin{lemma}
\label{lemma:upper}
    Suppose the condition $c$ is discrete, a  upper bound of \cref{eq:joint_kl} can be computed by:
    \begin{equation}
    \small
    \label{eq:upper}
    \begin{aligned}
    & \E_t\lambda_t\mathrm{KL}(p_\theta(\rvx_t,c) || p(\rvx_t, c)) \\
    & \leq \lambda_t\E_{p_\theta(\rvx_t|c)p(c),t} [-\log p(c|\rvx_t)p_\phi(\rvx_t) + \log p_\theta(\rvx_t)]\\
\end{aligned}
\end{equation}
\end{lemma}
See proof in the \cref{app:derivation}. The gradient of the upper bound for learning conditional generator $p_\theta(\rvx|c)$ can be computed as follows:
\begin{equation}
\small
\begin{aligned}
    \mathrm{Grad}(\theta) = - \alpha_t \E_{p_\theta(\rvx_t|c)p(c),t}\lambda_t[ & \underbrace{\nabla_{\rvx_t} \log p(c|\rvx_t)}_\text{condition learning} \\
    & + \underbrace{\nabla_{\rvx_t} \log \frac{p_{\phi}(\rvx_t)}{p_{\theta}(\rvx_t)}}_\text{fidelity learning}] \frac{\partial \rvx}{\partial \theta}.
\end{aligned}
\end{equation}
We can approximates $\nabla_{\rvx_t} \log p_{\theta}(\rvx_t)$ by a score model $s_\psi(\rvx_t,t)$, which is readily learnable through initialization via $s_\phi(\rvx_t,t)$. Following GAN's tradition~\cite{goodfellow2014generative}, we call $s_\phi$ as real score and $s_\psi$ as fake score.

This upper bound naturally decomposes into two distinct learning components: conditional alignment, and generation fidelity. This decomposition contrasts with VSD and diff-instruct frameworks, where the fake score serves multiple purposes in both conditional learning and fidelity learning, potentially compromising effectiveness. Our approach reduces the burden on the fake score while eliminating the requirement for teachers to understand training conditions.

\spara{Learning Fake Score}  We employ an auxiliary diffusion model $s_\psi$ to model $\nabla_{\rvx_t}\log p_\theta(\rvx_t)$. The fake score can be efficiently learned through denoising:
\begin{equation}
    \E_{t,p(\epsilon),p_\theta(\rvx)} || \epsilon_\psi(\rvx_t,t) - \epsilon||_2^2,
\end{equation}
where $\rvx_t = \alpha_t \rvx + \sigma_t \epsilon$ and $p(\epsilon)$ is the standard Gaussian distribution. After trained, we have $\nabla_{\rvx_t} \log p_{\theta,t}(\rvx_t) \approx s_\psi(\rvx_t,t) = -\frac{\epsilon_\psi(\rvx_t,t)}{\sigma_t}$.

\spara{Modeling the $\log p(c|\rvx_t)$} Unfortunately, conditional alignment density is defined in terms of clean samples in most cases, and rarely in terms of noisy samples. Hence, we need to find a way to approximate it. The $\log p(c|\rvx_t)$ can be modeled as follows:
\begin{equation}
\begin{aligned}
    p(c|\rvx_t) & \triangleq \int p(c|\rvx_t,\rvx)p(\rvx|\rvx_t) d\rvx \\
                & = \int p(c|\rvx)p(\rvx|\rvx_t) d\rvx
\end{aligned}
\end{equation}
where we substitute $p(c|\rvx_t,\rvx)$ with $p(c|\rvx)$, since the condition $c$ is fully relied on $\rvx$.
The remaining challenge is how to model $p(\rvx|\rvx_t)$. We propose to parameterize $p(\rvx|\rvx_t)$ using an \textit{implicit generator}. An easy way can be directly parameterizing it with the fake score. However, the distribution is defined over clean samples, and using a fake score can not estimate the distribution accurately. Hence, we propose parameterizing it using a consistency model, which can be efficiently trained upon the fake score through LoRA fine-tuning.
We note that modeling $p(c|\rvx_t)$ via consistency models by itself is interesting, how to integrate this technique with training-free controllable generation will be promising future work,  however, it is beyond the scope of this work.

\spara{Learning the $p(\rvx|\rvx_t)$} The $p(\rvx|\rvx_t)$ is modeled by a consistency model, we suggest learning the model by inserting a LoRA over the fake score for efficiency. Specifically, the consistency model~\cite{song2023consistency} can be efficiently trained through:
\begin{equation}
\small
\begin{aligned}
    &\min_{\beta} \E_{k,p(\epsilon),p_\theta(\rvx)} || f_{\psi,\beta}(\rvx_{t_k},t_k) - \mathrm{sg}(f_{\psi,\beta}(\rvx_{t_{k-1}},t_{k-1}))||_2^2, \\
    & f_{\psi,\beta}(\rvx_t,t) \triangleq 
\begin{cases} 
\rvx_t, & \text{if } t =  0 \\
\frac{\rvx_t - \sigma_t\epsilon_{\psi,\beta}(\rvx_t,t)}{\alpha_t}, & \text{if } t > 0 
\end{cases}\\
\end{aligned}
\end{equation}
where $t_k > t_{k-1}$, $\beta$ denotes the parameters of a lightweight LoRA, and $\mathrm{sg}(\cdot)$ denotes the stop-gradient operator.

\spara{Remark} Our learning framework supports universal guidance, as demonstrated through the following instances.

\subsection{Learning Better Aligned One-Step Generator}
\spara{Human Feedback Integration}
We demonstrate that Human-Feedback Learning (HFL) can be seamlessly integrated into this framework. Specifically, we introduce ``human-preferred images" as a single conditioning factor. Since we are dealing with only one condition, it is unnecessary to inject it into the desired generator. The learning gradient can then be expressed as:

\begin{equation}
\small
\begin{aligned}
    \mathrm{Grad}(\theta) = - \alpha_t \E_{p_\theta(\rvx_t|c)p(c),t}\lambda_t[ & \nabla_{\rvx_t} r(\rvx_t) \\
    & + \nabla_{\rvx_t} \log \frac{p_{\phi}(\rvx_t)}{p_{\theta}(\rvx_t)}] \frac{\partial \rvx}{\partial \theta},
\end{aligned}
\end{equation}
where $\nabla_{\rvx_t} r(\rvx_t)$ models the gradient of log probability of an image being human-preferred, formally defined as $\nabla_{\rvx_t} \log p(\text{``Human-preferred images"}|\rvx_t) \triangleq \nabla_{\rvx_t} r(\rvx_t) $. 

\spara{Decoupled CFG}
Our framework separates the learning of condition and fidelity into two distinct components. When the conditional probability $p(c|\rvx_t)$ represents text-image alignment, its gradient can be computed using Classifier-Free Guidance (CFG). While previous approaches~\cite{yin2023one} have incorporated CFG during distillation, however, their CFG is coupled with the real score. In contrast, our framework explicitly leverages CFG for conditional learning. This key difference allows us to compute CFG using a diffusion model different from the one used for real score computation, enabling the use of more sophisticated diffusion models to guide text-image alignment.

\subsection{Learning One-Step Generator with Additional Control}
\label{sec:one_step_control}

\spara{Controllable Generation}
To incorporate additional control capabilities similar to ControlNet~\cite{zhang2023adding}, we parameterize the student model as a diffusion model with an associated ControlNet. The generator's learning objective becomes:
\begin{equation}
\label{eq:learn_ctrlnet}
    \min_{\theta,\beta} -\E_{p_{\theta,\beta}(\rvx_t|c)p(c),t} \lambda_t [ \log p(c|\rvx_t) + \log \frac{ p_{\phi,t}(\rvx_t)}{p_{\psi,t}(\rvx_t)} ],
\end{equation}
where $\beta$ denotes the additional parameters for ControlNet. 
This formulation enables asymmetric capability development - the student model can learn conditional generation tasks beyond the teacher's abilities.

\begin{figure}[!t]
    \centering
    \includegraphics[width=1\linewidth]{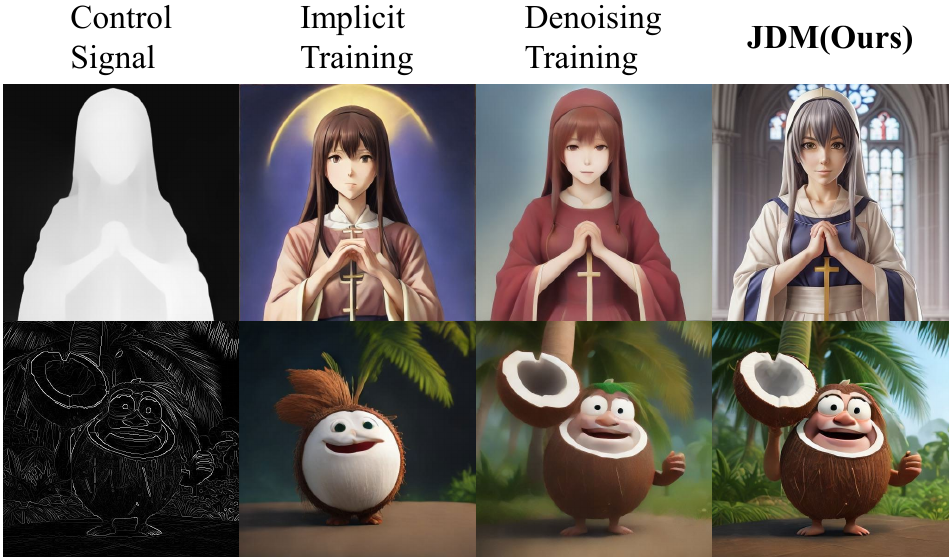}
    \vspace{-5mm}
    \caption{The qualitative comparison of the proposed method and potential baselines in one-step controllable generation.}
    \vspace{-4mm}
    \label{fig:ablation}
\end{figure}

To the best of our knowledge, we are the first to explore learning additional control one-step generator by score distillation in an asymmetric form. To highlight the effectiveness of our proposed formulation, we discuss two potential competitive variants for learning one-step generators with additional control conditions. 
1) \textbf{Train ControlNet via Denoising:} Given a pre-trained diffusion model $\epsilon_\phi(\rvx_t, t)$, a ControlNet parameterized by $\beta$ can be trained by denoising score matching loss. Hence a naive idea is training the ControlNet for one-step generation using denoising loss too. Empirically, we find this approach can transfer the control well but generate blurry images (see \cref{tab:ablation} and \cref{fig:ablation}); 
2) \textbf{Implicitly Train ControlNet via Score Distillation} Training ControlNet for Diffusion Models is straightforward by directly using the original diffusion loss. Hence, a natural idea is to follow the training of ControlNet for Diffusion Models and directly inject the condition via ControlNet with the original VSD loss. This approach shares a high-level idea with training ControlNet for diffusion models: maintaining the original training loss while only adding ControlNet to the generator for condition injection. However, we empirically find this approach can generate high-quality images but ignores the control signal (see \cref{tab:ablation} and \cref{fig:ablation}).

\subsubsection{Shared One-Step Generator Between Different Additional Control}
While \cref{eq:learn_ctrlnet} establishes the joint training of One-Step Generator $G_\theta$ and ControlNet $\phi$, this approach necessitates retraining the entire One-Step Generator for each new control condition. This requirement leads to inefficient use of computational resources and storage space, which limits practical applications.

A simple solution would be to first train a one-step generator without control using diff-instruct~\cite{luo2023diff}, then train ControlNet to incorporate additional control into $G_\theta$. However, previous work has shown that one-step generators trained in this manner often suffer from mode collapse, making it difficult to accommodate additional control signals. One potential approach is to use teacher diffusion to generate millions of noise-image pairs, then add ODE regression loss for a one-step generator~\cite{yin2023one}. However, this method is extremely computationally inefficient and impractical for real-world applications.

To address these limitations, we propose a novel two-phase warm-up training strategy:
\noindent\textbf{1) Initial Phase:} Joint training of $G_\theta$ and ControlNet for a primary condition;
\noindent\textbf{2) Extension Phase:} Fixing $G_\theta$ while training only ControlNet for subsequent conditions. 
This approach leverages the joint KL divergence to regularize $G_\theta$ during initial training, resulting in better accommodating the condition and avoiding mode collapse. The resulting well-trained $G_\theta$ can then effectively incorporate other forms of control.
\vspace{-2mm}
\section{Experiments}
\vspace{-1mm}

\begin{table*}[t]
        \centering
        \caption{Comparison of machine metrics of different methods across tasks.}
        \label{tab:main_control}
        \resizebox{\linewidth}{!}{
        \begin{tabular}{l|c|cc|cc|cc|cc|cc}
        \toprule
        \multirow{2}{*}{Method} & \multirow{2}{*}{NFE$\downarrow$} & \multicolumn{2}{|c|}{Canny} & \multicolumn{2}{|c|}{HED} & \multicolumn{2}{|c|}{Depth} & \multicolumn{2}{|c|}{8$\times$ Super Resolution} & \multicolumn{2}{|c|}{Avg}  \\
        ~ & ~ & FID$\downarrow$ & Consistency$\downarrow$ & FID$\downarrow$ & Consistency$\downarrow$ & FID$\downarrow$ & Consistency$\downarrow$ & FID$\downarrow$ & Consistency$\downarrow$ & FID$\downarrow$ & Consistency$\downarrow$ \\
        \midrule
        ControlNet & 50 & 14.48 & 0.113 & 19.21 & 0.101 & \textbf{15.25} & 0.093 & 11.93 & 0.065 & 15.21 & 0.093 \\
        \midrule
        DI + ControlNet & 1 & 22.74 & 0.141 & 28.04 & 0.113 & 22.49 & 0.097 & 15.57 & 0.126 & 22.21 & 0.120 \\
        \textbf{\shortname (Ours)} & 1 & \textbf{13.07} & \textbf{0.102} & \textbf{16.51} & \textbf{0.045} & 16.24 & \textbf{0.081 }& 12.48 & \textbf{0.053} & \textbf{14.58} & \textbf{0.071}\\
        \bottomrule
        \end{tabular}
        }
\end{table*}

\begin{figure*}[t]
    \centering
    \includegraphics[width=1\linewidth]{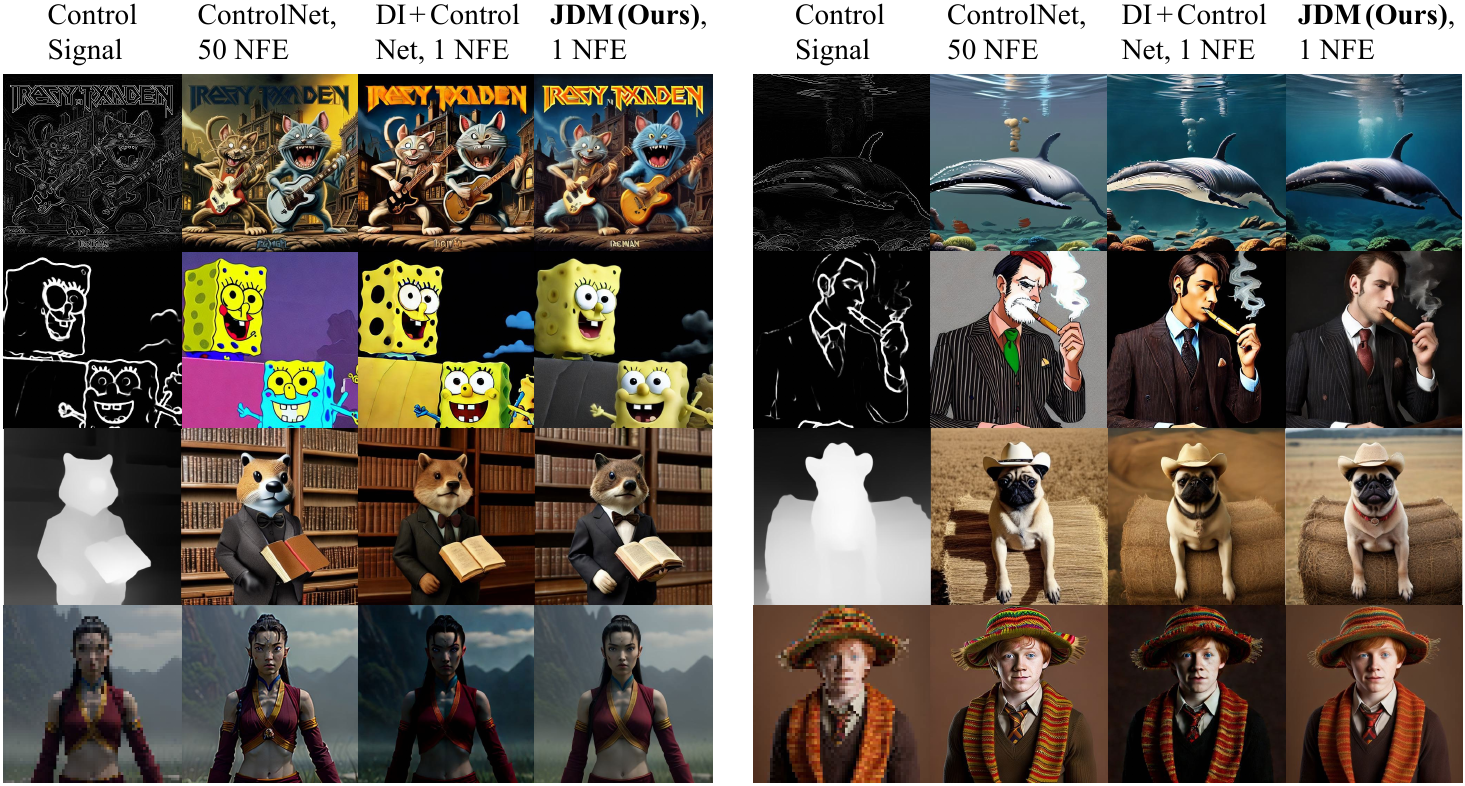} 
    \vspace{-4mm}
    \caption{Qualitative comparisons on controllable generation across different control signals against competing methods.}
    \label{fig:comparisons_control}
\end{figure*}

\subsection{Controllable Generation}
\spara{Experiment Setting} All the models are trained in an internally collected dataset. We use Stable Diffusion 1.5~\cite{rombach2022high} as the frozen teacher model and initialization for the fake score model and student. The ControlNet is initialized in the way introduced in its original paper~\cite{zhang2023adding}. We utilize four conditions for evaluating our method in one-step controllable generation as follows: Canny~\cite{canny1986computational}, Hed~\cite{xie2015holistically}, Depth map, and lower resolution samples. Details for these conditions can be found in the \cref{app:condition_detail}.

\spara{Evaluation Metric} We select the FID~\cite{heusel2017gans} to measure the image quality and consistency to measure the controllability. In particular, the FID is calculated between generated images by SD without additional control and generated images based on additional conditions. The consistency is defined by $\mathrm{Consistecny} = || h(\rvx) - c||_1$, where $c$ is the condition, and $h(\cdot)$ is the function used for obtaining condition. We also report the number of function evaluations (NFE) required for generating an image for comparing efficiency.

\spara{Quantitative Results} We conduct comprehensive evaluations comparing our proposed approach against two baseline methods: (1) the standard diffusion model (DM) with ControlNet and (2) pre-trained one-step generator by diff-instruct integrated with DM's ControlNet. The quantitative results are summarized in \cref{tab:main_control}, where we evaluate both image quality (FID) and condition consistency across multiple control tasks. Our results reveal several key findings:
1) The proposed method achieves a significant reduction in the number of function evaluations (NFEs) from 50 to 1, while maintaining or improving performance metrics. Specifically, our method demonstrates superior FID scores and consistency measures across various conditioning tasks, indicating both better image quality and more precise condition adherence.
2) Our shared one-step generator architecture exhibits substantial improvements over the naive integration of a pre-trained one-step generator, validating the effectiveness of our unified training strategy. This is evidenced by consistent performance gains across all evaluated metrics and tasks.
3) While it is technically feasible to directly combine DM's ControlNet with a pre-trained one-step generator, this approach yields substantially inferior results. This observation underscores the importance of our proposed strategy tailored for training one-step generators with additional control. Overall, these results demonstrate that our method successfully achieves better trade-off between efficiency and sample quality in controlled image generation, achieving state-of-the-art performance with significantly reduced computational overhead

\spara{Qualitative Comparison} We present the qualitative comparison in \cref{fig:comparisons_control}. We can observe that DM's ControlNet provides high-level control for one-step generators. However, this approach often produces lower-quality images. Our customized one-step generator training with additional control can generate much higher-quality images. This validates the effectiveness of our proposed method and indicates that we can teach students conditions unknown to teachers by minimizing the upper bound of joint KL divergence.

\subsection{Comparison to Potential Baselines}
We conduct comprehensive studies to validate our design choices by comparing with two potential baseline approaches as discussed in \cref{sec:one_step_control}: 1)training ControlNet via denoising loss; 2) implicitly training ControlNet via score distillation. The quantitative results are shown in  \cref{tab:main_control}. The quantitative results are shown in \cref{fig:ablation}.

\begin{figure*}[t]
    \centering
    \includegraphics[width=.95\linewidth]{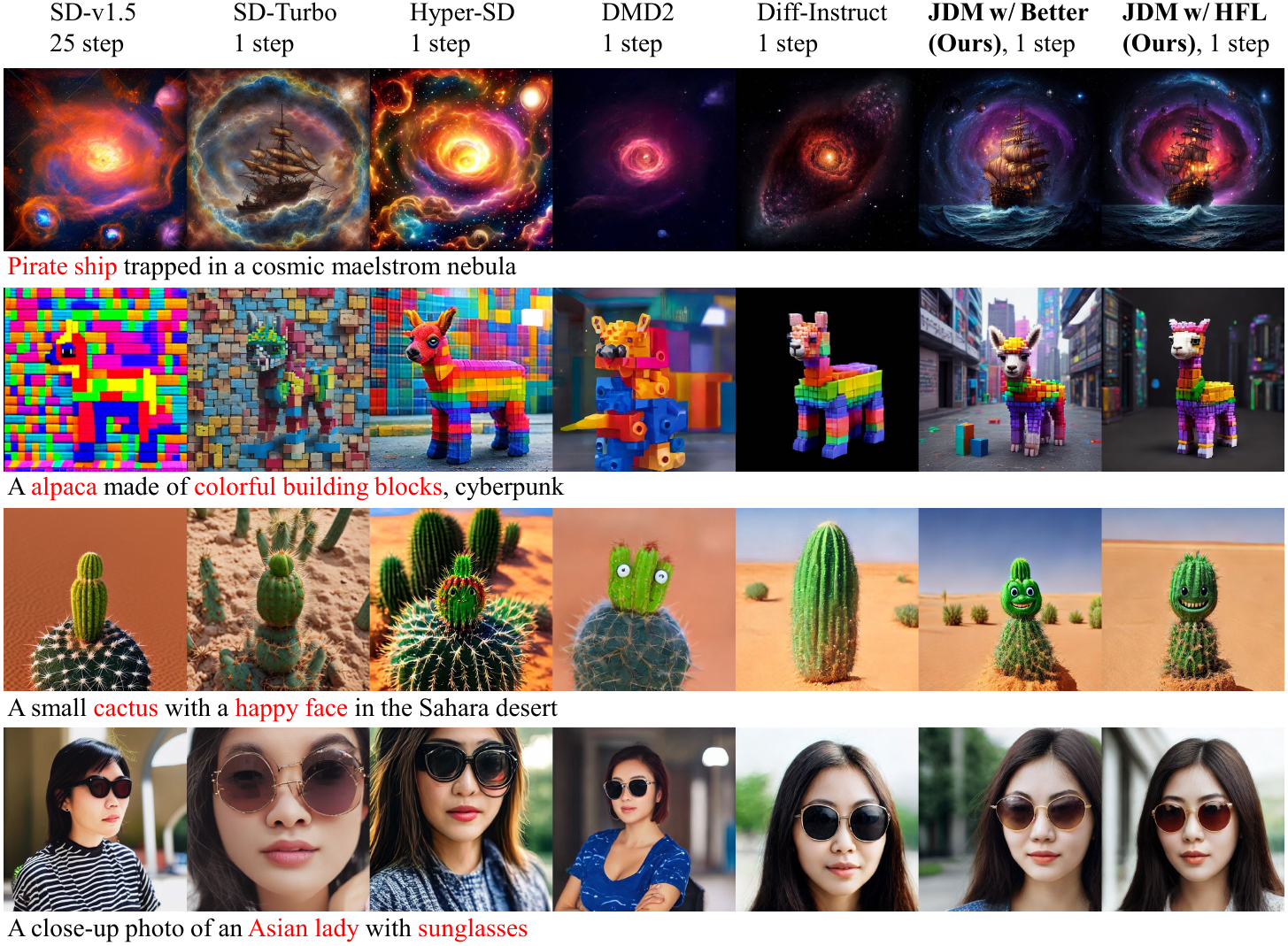}
    \vspace{-3mm}
    \caption{Qualitative comparisons on text-to-image generation across different control signals against competing methods.}
    \label{fig:t2i}
    \vspace{-5mm}
\end{figure*}

\spara{Baseline Details} We introduce the training of two potential baselines as follows:
\begin{itemize}
    \item  \textbf{Train ControlNet via Denoising:}  Given a pre-trained one-step diffusion model $G_\theta$, the ControlNet is trained by denoising:
\begin{equation}
    \small
     L(\beta) = \E_{\rvx, \epsilon}|| G_{\{\phi, \beta\}}(\rvx_T, c) - \rvx ||_2^2,
\end{equation}
where $T$ is the terminal step chosen in the one-step generator.

\item \textbf{Implicitly Train ControlNet via score distillation:} The one-step student with additional control is directly trained by the original VSD loss as following:
\begin{equation}
\small
\label{eq:implicit_ctrl}
    \mathrm{KL}(\int_c q_{\theta,\beta}(\rvx|c)p(c)dc || p_\phi(\rvx) ).
\end{equation}
The condition injection is implicitly trained via reverse KL divergence. 
\end{itemize}

\spara{Comparing to Train ControlNet via Denoising} The baseline of training ControlNet through DSM loss achieves reasonable consistency scores (0.109 for Canny and 0.094 for Depth) but suffers from significantly degraded image quality, as reflected by the worse FID scores (28.83 for Canny and 35.14 for Depth). This reveals that this approach tends to generate blurry images while maintaining decent control. The degradation in image quality can be attributed to the mismatch between the denoising objective and the reverse KL divergence, as the model is forced to learn denoising behavior that may not be optimal for direct generation.

\spara{Comparing to Implicitly Train ControlNet via score distillation} The implicit training baseline through score distillation shows better FID scores (22.87 for Canny and 22.34 for Depth) compared to the DSM approach, indicating its capability to generate higher quality images. However, it demonstrates worse consistency scores (0.151 for Canny and 0.150 for Depth), suggesting that the control signals are not effectively incorporated. This indicates that simply maintaining the original VSD loss while adding ControlNet may lead to the model ignoring the control conditions.

\begin{table}[t]
        \centering
        \caption{Comparison of machine metrics of different methods across tasks.}
        \vspace{-4mm}
        \label{tab:ablation}
        \resizebox{\linewidth}{!}{
        \begin{tabular}{l|c|cc|cc}
        \toprule
        \multirow{2}{*}{Method} & \multirow{2}{*}{NFE$\downarrow$} & \multicolumn{2}{|c|}{Canny} & \multicolumn{2}{|c|}{Depth}\\
        ~ & ~ & FID$\downarrow$ & Consistency$\downarrow$ & FID$\downarrow$ & Consistency$\downarrow$ \\
        \midrule
        \textbf{\shortname (Ours)} & 1 & \textbf{13.07} & \textbf{0.102} &  \textbf{16.24} & \textbf{0.081} \\
        \midrule
        w/ warm-up shared UNet & 1 & 14.35 & 0.122 & 16.71 & 0.093 \\
        w/o modeling $p(c|\rvx_t)$ & 1 & 14.89 & 0.113 & 18.56 & 0.085 \\
        \midrule
        \midrule
         Implicit Training & 1 & 22.87 & 0.151  & 22.34 & 0.150  \\
         DI w/ Denoising & 1 & 28.83 & 0.109 & 35.14 & 0.094  \\
        \bottomrule
        \end{tabular}
        \vspace{-4mm}
        }
\end{table}

\begin{table*}[t]
        \centering
        \caption{Comparison of machine metrics on text-to-to-image generation across state-of-the-art methods. HFL denotes human feedback learning which might hack the machine metrics. We highlight the \textbf{best} and \textit{second best} among distillation methods.}
        \label{tab:main_t2i}
        \vspace{-3mm}
        \resizebox{1\linewidth}{!}{
        \begin{tabular}{lcccccccccc}
        \toprule
             \multirow{2}{*}{Model} & \multirow{2}{*}{Backbone} & \multirow{2}{*}{HFL} & \multirow{2}{*}{Steps} & \multicolumn{5}{|c|}{HPS$\uparrow$} & \multirow{2}{*}{Aes$\uparrow$}  & \multirow{2}{*}{CS$\uparrow$} \\
              &  &  &  & \multicolumn{1}{|c}{Animation} & Concept-Art & Painting & Photo & \multicolumn{1}{c|}{Average} &  &\\
        \midrule
             \rowcolor{gray!20}
             Base Model & SD-v1.5 & No & 25 & 26.29 & 24.85 & 24.87 & 26.01 & 25.50 & 5.49 & 33.03 \\
             \rowcolor{gray!20}
             Base Model & SD-v2.1 & No & 25 & 27.82 & 27.14 & 27.17 & 28.17 & 27.58 & 5.66 & 33.46 \\
             SD Turbo~\citep{sauer2023adversarial} & SD 2.1 & No & 1 & 28.30 & 26.92 & 26.43 & 25.54 & 26.80 & 5.31 & 32.21 \\
             InstaFlow~\citep{liu2023insta} &  SD-v1.5  & No & 1  & 23.17 & 23.04 &22.73 &22.97 & 22.98 & 5.25  & 31.97 \\
             TCD~\citep{tcd} &  SD-v1.5 & No & 4    & 23.14 & 21.11 & 21.08 & 23.62 & 22.24 & 5.43 & 29.07 \\
             LCM-dreamshaper~\citep{luo2023lcmlora} &  SD-v1.5 & No & 4    & 26.51 & 26.40 & 25.96 & 24.32 & 25.80 & 5.94 & 31.55 \\
             PeRFlow~\citep{yan2024perflow} &  SD-v1.5  & No & 4  & 22.79 & 22.17 & 21.28 & 23.50 & 22.43 & 5.35 & 30.77 \\
             DMD2~\cite{dmd2} &  SD-v1.5 & No & 1    & 24.17 & 22.68 & 22.97 & 24.30 & 23.53 & 5.82 & 30.92 \\
             Diff Instruct~\cite{luo2023diff} &  SD-v1.5 & No & 1    & 27.32 & 26.15 & 26.41 & 25.50 &  26.35 & 5.71 & 32.08 \\
             Hyper-SD~\citep{ren2024hypersd} &  SD-v1.5  & Yes & 1  & 28.65 & 28.16 & 28.41 & 26.90 & 28.01 & 5.64 & 30.87 \\
             \midrule
             \textbf{\shortname w/ HFL (Ours)} &  SD-v1.5 & Yes & 1    & \textit{30.16} & \textit{29.17} & \textit{30.14} & \textit{28.35} &  \textit{29.46} & \textit{5.89} & \textit{33.75} \\
             \textbf{\shortname w/ better CFG (Ours)} &  SD-v1.5 & No & 1    & \textbf{30.56} & \textbf{29.46} & \textbf{30.38} & \textbf{28.59} &  \textbf{29.75} & \textbf{5.90} & \textbf{33.97} \\
        \bottomrule
        \end{tabular}
                }
\end{table*}

\subsection{Other Application in Text-to-Image Generation}

\spara{Evaluation} We employs multiple metrics to assess different aspects of generation quality:  Aesthetic Score (AeS)~\citep{schuhmann2022laion} evaluates the image quality; CLIP Score (CS) measures the text-to-image alignment; Human Preference Score (HPS) v2.1~\citep{wu2023human} correlates strongly with the human preference, capturing both image-text alignment and aesthetic quality. 

\spara{Baseline Models} We conduct our experiments on SD-v1.5. 
We perform \textit{human feedback learning} by ImageReward~\cite{xu2024imagereward}. To ensure fairness in evaluation, we do not report this metric
We perform \textit{decoupled CFG} by using SD-v2.1 in distilling SD-v1.5.
We mainly compare our model against the open-source state-of-the-art (SOTA) models, e.g., LCM~\cite{luo2023latent},  Hyper-SD~\citep{ren2024hypersd}, and DMD2~\cite{dmd2}.

\spara{Quantitative Results} Table~\ref{tab:main_t2i} presents a comprehensive comparison of our method against existing approaches. Our one-step generator, enhanced with SD-v2.1-based CFG and human feedback learning, demonstrates superior performance across all evaluation metrics. Notably, our method significantly outperforms the direct baselines such as Diff-Instruct~\cite{luo2023diff} and DMD2~\cite{dmd2}. An intriguing observation is that the variant utilizing better CFG achieves even better metrics compared to the HFL variant. This unexpected finding suggests a promising research direction: the potential benefits of leveraging multiple teacher diffusion models for student model distillation, which merits further investigation in future work.

\spara{Qualitative Comparison} 
Since using HFL may lead to the hack of machine metrics, we further conducted qualitative comparisons. We qualitatively compared our w/ HFL and our w/ better CFG with the most competitive baseline, as shown in \cref{fig:t2i}. Our method demonstrates significantly better visual quality and text-image alignment. Notably, while Hyper-SD also employs HFL, their HFL is performed separately, whereas ours is conducted collaboratively with one-step learning, allowing fake scores to participate and eliminate artifacts caused by reward maximization. As shown in \cref{fig:t2i}, Hyper-SD exhibits noticeable artifacts, likely due to their flawed HFL approach.

\begin{figure}[!t]
    \centering
    \includegraphics[width=1\linewidth]{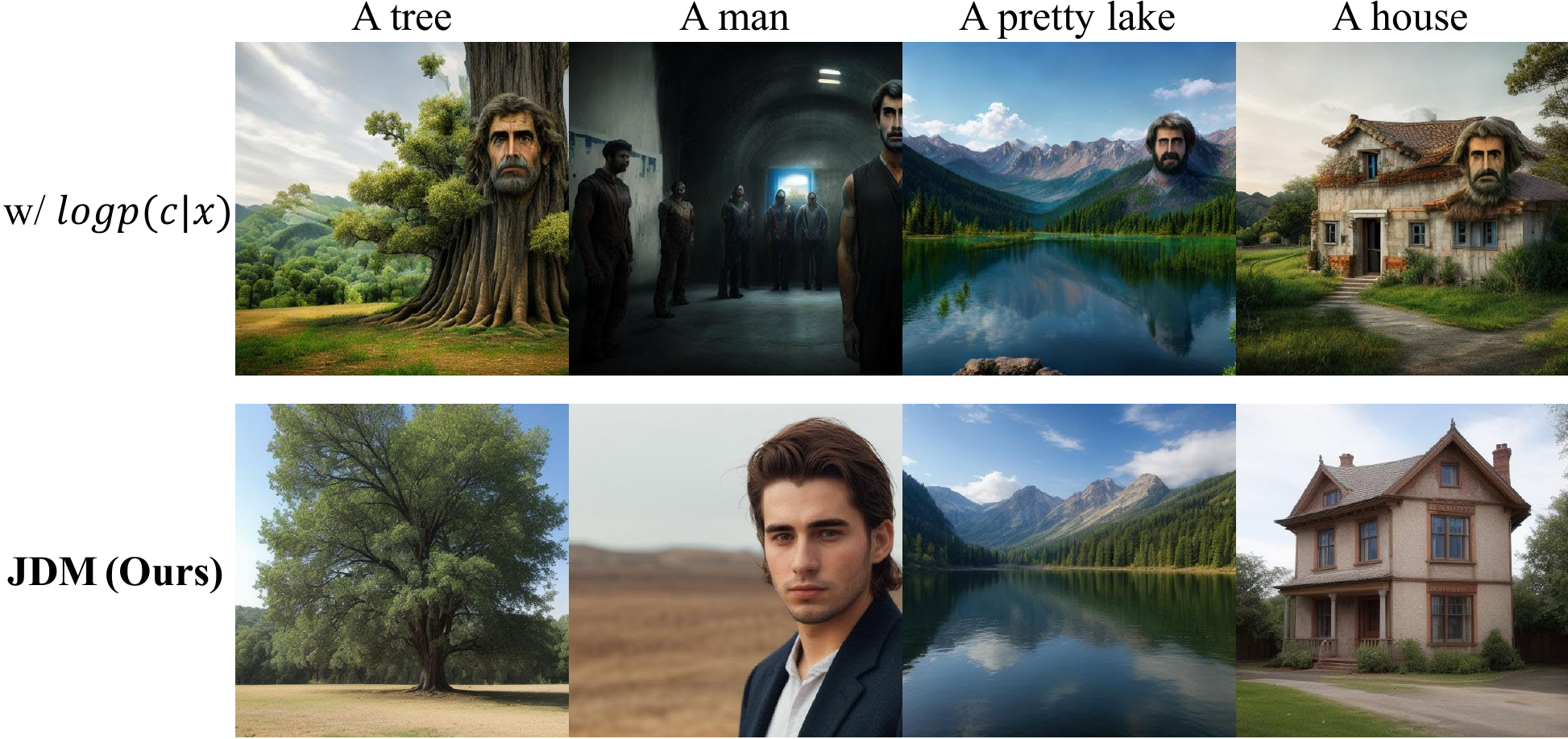}
    \vspace{-5mm}
    \caption{We compare JDM with the variant directly using $\log p(c|\rvx)$. It is clear that the variant suffers  from \textit{artifacts}.}
    \label{fig:rlhf_ablation}
\end{figure}

\subsection{Additional Ablation Study}

\paragraph{Effect of Warm-up Shared UNet} The proposed two-phase warm-up training strategy is crucial for learning a shared one-step UNet among different controls. Here, we compare our strategy to naively using pre-trained DI as shared UNet. The results are shown in \cref{tab:ablation}. Without the warm-up strategy, the performance will degrade severely, especially regarding the consistency metric. This indicates our warm-up strategy can help in learning a One-step UNet that can handle multiple conditions well.

\paragraph{Effect of Modeling $\log p(c|\rvx_t)$} 
Modeling $\log p(c|\rvx_t)$ not only makes the formulation more elegant but also ensures that the information from the gradients of condition learning and fidelity pertains to the same diffused $\rvx_t$. If we were to simply substitute $\log p(c|\rvx_t)$ with $\log p(c|\rvx_0)$, the gradient information would relate separately to $\rvx_t$ and $\rvx_0$, resulting in less stable learning. As demonstrated in \cref{tab:ablation}, omitting the modeling of $\log p(c|\rvx_t)$ leads to a significant drop in performance, particularly concerning the FID. This highlights the importance of modeling $\log p(c|\rvx_t)$ for achieving high-quality controllable one-step generation. Furthermore, modeling $\log p(c|\rvx_t)$ isis more robust than using $\log p(c|\rvx_0)$. In HFL learning scenarios, directly modeling $\log p(c|\rvx_0)$ can result in noticeable artifacts caused by \textit{reward hacking}, as illustrated in \cref{fig:rlhf_ablation}.

\section{Conclusion}

In this work, we propose JDM for adding new control unknown to teacher DMs into one-step student. Our method minimizes the upper bound of reverse KL divergence between image-condition joint distributions. This approach decouples fidelity and condition learning, allowing the one-step student to handle controls unknown to the teacher. Extensive experiments show that JDM outperforms multi-step controllable DMs by one-step, while achieving SOTA performance in one-step text-to-image synthesis by the integration of decoupled CFG or human feedback learning.

{
    \small
    \bibliographystyle{ieeenat_fullname}
    \bibliography{main}
}

\appendix
\clearpage
\setcounter{page}{1}
\maketitlesupplementary

\section{Proof of \cref{lemma:upper}}
\label{app:derivation}
Since we assume the condition $c$ is discrete, its entropy $\mathcal{H}(c)$ and conditional entropy $\mathcal{H}(c|\rvx_t)$ would be non-negative. Combine $\mathcal{H}(x_t,c) = \mathcal{H}(x_t) + \mathcal{H}(c|x_t)$, we have:
\begin{equation}
\label{eq:upper_entropy}
\begin{aligned}
    \mathcal{H}(x_t,c) & = -\E_{p_\theta(\rvx_t,c)}\log p_\theta(\rvx_t,c) \\
    & \geq\mathcal{H}(x_t) = -\E_{p_\theta(\rvx_t)}\log p_\theta(\rvx_t).
\end{aligned}
\end{equation}
By substituting \cref{eq:upper_entropy} into the integral joint KL divergence $\E_t \lambda_t\mathrm{KL}(p_\theta(\rvx_t,c) || p(\rvx_t, c))$, we have:
\begin{equation}
\small
\begin{aligned}
    & \E_t\lambda_t\mathrm{KL}(p_\theta(\rvx_t,c) || p(\rvx_t, c)) \\
    & =  - \lambda_t\E_{p_\theta(\rvx_t|c) p(c),t} \log p(c|\rvx_t)p_\phi(\rvx_t) + \lambda_t\E_{p_\theta(\rvx_t,c)} \log p_\theta(\rvx_t,c)\\ 
    & \leq - \lambda_t\E_{p_\theta(\rvx_t|c)p(c),t} \log p(c|\rvx_t)p_\phi(\rvx_t) + \lambda_t\E_{p_\theta(\rvx_t)} \log p_\theta(\rvx_t)\\
    & = \lambda_t\E_{p_\theta(\rvx_t|c)p(c),t} [-\log p(c|\rvx_t)p_\phi(\rvx_t) + \log p_\theta(\rvx_t)]
\end{aligned}
\end{equation}
This completes the proof.

\section{Details of conditions}
\label{app:condition_detail}
\begin{itemize}
    \item Canny: a canny edge detector~\cite{canny1986computational} is employed to generate canny edges;
    \item Hed: a holistically-nested edge detection model is utilized for the purpose;
    \item Depthmap: we employ the Midas~\cite{ranftl2020towards} for depth estimation;
    \item Super-resolution: we use the nearest kernel to downscale the images by a factor of $8$ as the condition.
\end{itemize}

\end{document}